%% file: main.tex
\pdfoutput=1

\documentclass[11pt]{article}

\usepackage{acl}

\usepackage{times}
\usepackage{latexsym}
\usepackage{graphicx}
\usepackage{subcaption}
\usepackage{microtype}
\usepackage{array}
\usepackage{pifont}
\usepackage{tabularx}
\usepackage{adjustbox}
\usepackage{multirow}
\usepackage{multicol}
\usepackage{enumitem}
\usepackage{xspace}
\usepackage{tcolorbox}
\usepackage{booktabs,amsfonts,dcolumn}
\usepackage{hyperref}
\usepackage{url}
\usepackage{amsmath,amsthm,amsfonts,amssymb,bm,stmaryrd}
\usepackage[noorphans,vskip=0.75ex,leftmargin=2ex]{quoting}
\usepackage{algorithm,algpseudocode}
\usepackage{makecell}
\usepackage{graphicx}
\usepackage{xcolor,colortbl}
\usepackage{color,soul}

\usepackage[T1]{fontenc}

\usepackage[utf8]{inputenc}

\usepackage{microtype}

\input{header}

\title{Ditch the Gold Standard: \\ Re-evaluating Conversational Question Answering}

\author{Huihan Li \\
  Affiliation / Address line 1 \\
  \texttt{huihanl@princeton.edu} \\\And
  Tianyu Gao \\
  Affiliation / Address line 1 \\
  \texttt{tianyug@princeton.edu} \\ \And
  Manan Goenka \\
  Affiliation / Address line 1 \\
  \texttt{mgoenka@princeton.edu} \\
  Danqi Chen \\
  Affiliation / Address line 1 \\
  \texttt{danqic@cs.princeton.edu} \\
  }

  \author{Huihan Li$^{*}$ \quad Tianyu Gao$^{*}$ \quad Manan Goenka \quad Danqi Chen \\
  Department of Computer Science, Princeton University\\
  \ttt{\{huihanl,tianyug,mgoenka,danqic\}@princeton.edu}\\
  }

\date{}

\begin{document}
\maketitle
\renewcommand{\thefootnote}{\fnsymbol{footnote}}
\footnotetext[1]{The first two authors contributed equally.}
\renewcommand{\thefootnote}{\arabic{footnote}}
\begin{abstract}

Conversational question answering aims to provide natural-language answers to users in information-seeking conversations.
Existing  {\convqa{}} benchmarks compare models with pre-collected human-human conversations, using ground-truth answers provided in conversational history.
It remains unclear whether we can rely on this static evaluation for model development and whether current systems can well generalize to real-world human-machine conversations.
In this work, we conduct the first large-scale human evaluation of state-of-the-art \convqa{} systems, where human evaluators converse with models and judge the correctness of their answers.
We find that the distribution of human-machine conversations differs drastically from that of human-human conversations,
and there is a disagreement between human and gold-history evaluation in terms of model ranking.
We further investigate how to improve automatic evaluations, and propose a question rewriting mechanism based on predicted history, which better correlates with human judgments.
Finally, we analyze the impact of various modeling strategies and discuss future directions towards building better conversational question answering systems.\footnote{Our data and code are publicly available at \url{https://github.com/princeton-nlp/EvalConvQA}.}
\end{abstract}

\input{sections/Intro}

\input{sections/Bg}
\input{sections/Human}
\input{sections/Goldvshuman}

\input{sections/Pred}
\input{sections/Automatic}
\input{sections/Experiment}

\input{sections/ModelAnalysis}
\input{sections/Related}
\input{sections/Conclusion}

\section*{Acknowledgements}
We thank Alexander Wettig and other members of the Princeton NLP group, 
and the anonymous reviewers  for  their valuable feedback.  
This research is supported by a Graduate Fellowship at Princeton University and the James Mi *91 Research Innovation Fund for Data Science.

\bibliography{custom}
\bibliographystyle{acl_natbib}

\clearpage

\appendix

\input{sections/Appendix}

\end{document}

%% file: header.tex
\providecommand{\bert}{BERT}
\providecommand{\ham}{HAM}
\providecommand{\excord}{ExCorD}
\providecommand{\graphflow}{GraphFlow}

\providecommand{\gold}{Auto-Gold}
\providecommand{\pred}{Auto-Pred}
\providecommand{\rewrite}{Auto-Rewrite}
\providecommand{\replace}{Auto-Replace}

\providecommand{\Replace}{Auto-Replace}

\providecommand{\unans}{\texttt{CANNOT} \texttt{ANSWER}}

\newcommand{\xmark}{\ding{55}}

\newcommand\ti[1]{\textit{#1}}

\newcommand\tf[1]{\textbf{#1}}
\newcommand\ttt[1]{\texttt{#1}}

\newcommand\human{\mathcal{H}}
\newcommand\model{\mathcal{M}}
\newcommand\Q{Q}
\newcommand\A{A}
\newcommand\Qg{Q^*}
\newcommand\Ag{A^*}
\newcommand\bg{BG}

\newcommand\convqa{conversational QA}

\renewcommand{\paragraph}[1]{\vspace{0.2cm}\noindent\textbf{#1}}

\newcommand\gcolor{\color{blue}}

\definecolor{cid}{HTML}{efefef}

%% file: sections/Intro.tex

\section{Introduction}
\label{sec:intro}



Conversational question answering aims to build machines to answer questions in conversations and has the promise to revolutionize the way humans interact with machines for information seeking.
With recent development of large-scale datasets~\cite{choi2018quac,saeidi2018interpretation,reddy2019coqa,campos-etal-2020-doqa}, rapid progress has been made in better modeling of conversational QA systems.

\begin{figure*}[ht!]
        \centering
        \includegraphics[width=.98\linewidth]{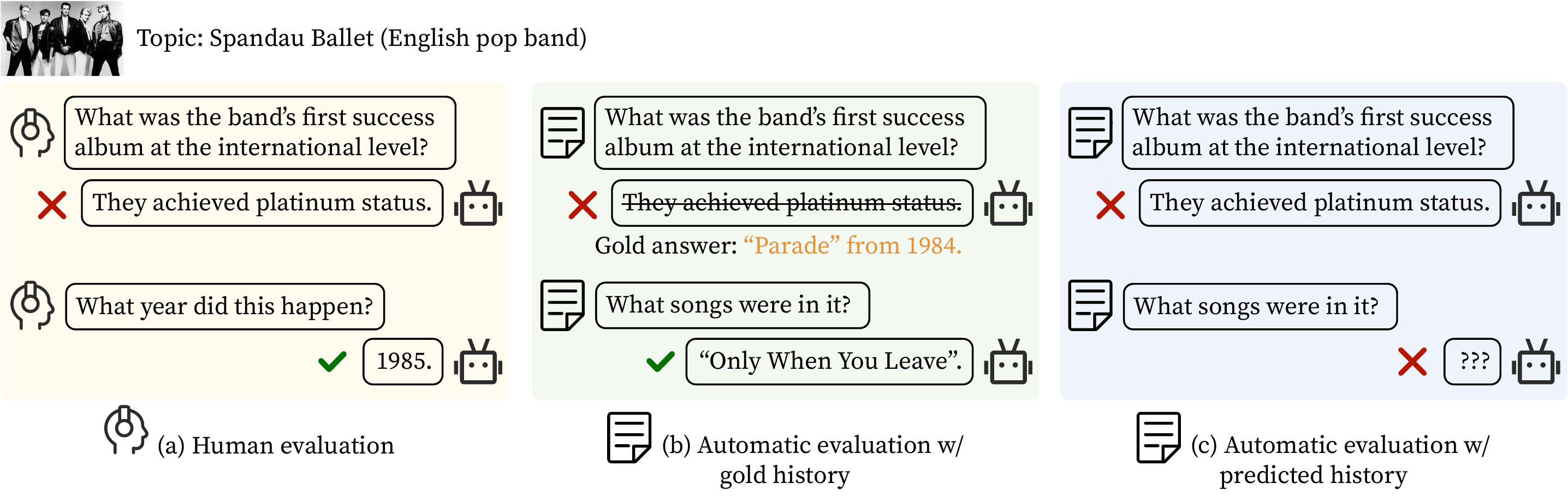}
    \caption{
        Examples of \ti{human} and \ti{automatic} evaluations with gold or predicted history.
        The model answers the first question incorrectly.
        (a) A human questioner asks the next question based on current predictions.
        (b) Automatic evaluation with gold history poses pre-collected questions with gold answers as conversational history, regardless of model predictions.
        (c) Using predicted history in automatic evaluation makes the next question invalid.
    }
    \vspace{-5pt}
    \label{fig:cqa}
\end{figure*}

Current  {\convqa{}} datasets are collected by crowdsourcing human-human conversations, where the questioner asks questions about a specific topic, and the answerer provides answers based on an evidence passage and the conversational history.
When evaluating {\convqa{}} systems, a set of held-out conversations are used for asking models questions in turn. Since the evaluation builds on pre-collected conversations, the \ti{gold  history} of the conversation is always provided, regardless of models' actual predictions (Figure~\ref{fig:cqa}(b)). Although current systems achieve near-human F1 scores on this static evaluation,
it is questionable whether this can faithfully reflect models' true performance in real-world applications.
To what extent do human-machine conversations deviate from human-human conversations? What will happen if models have no access to ground-truth answers in a conversation?

To answer these questions and better understand the performance of \convqa{} systems,
we carry out the first
large-scale human evaluation with four state-of-the-art models on the QuAC dataset~\cite{choi2018quac}
by having human evaluators converse with the models
and judge the correctness of their answers.
We collected 1,446 human-machine conversations in total, with 15,059 question-answer pairs.
Through careful analysis, we notice a significant distribution shift from human-human conversations and identify a clear inconsistency of model performance between current evaluation protocol and human judgements.

This finding motivates us to improve automatic evaluation such that it is better aligned with human evaluation.
\citet{mandya-etal-2020-history,siblini-etal-2021-towards} identify a similar issue in gold-history evaluation and propose to use models' own predictions for automatic evaluation.
However, predicted-history evaluation poses another challenge: since all the questions have been collected beforehand, using predicted history will invalidate some of the questions because of changes in the conversational history (see Figure~\ref{fig:cqa}(c) for an example).
Following this intuition,
we propose a \emph{question rewriting} mechanism, which automatically detects and rewrites invalid questions with predicted history (Figure~\ref{fig:rewrite}).
We use a coreference resolution model~\cite{lee2018higher} to detect inconsistency of conference in question text conditioned on predicted history and gold history, and then
rewrite those questions by substituting with correct mentions, so that the questions are resolvable in the predicted context. 
Compared to predicted-history evaluation, we find that incorporating this rewriting mechanism
 aligns better with human evaluation.


Finally, we also investigate the impact of different modeling strategies based on human evaluation. We find that both accurately detecting unanswerable questions and explicitly modeling question dependencies in conversations are crucial for model performance.
Equipped with all the insights, we discuss directions for {\convqa{}} modeling.
We release our human evaluation dataset and
hope that our findings can shed light on future development of better {\convqa{}}  systems.

%% file: sections/Bg.tex
\section{Preliminary}
\label{sec:bg}
\subsection{Evaluation of conversational QA}


Evaluation of conversational QA in real-world consists of three components: an evidence passage $P$, 
a (human) questioner $\human$ that has no access to $P$,\footnote{Existing conversational QA datasets make different assumptions: For example, QuAC~\cite{choi2018quac} assumes no access but CoQA assumes the questioner to have access.} 
and a model $\model$ that has access to $P$.
The questioner asks questions about $P$ and the model answers them based on $P$ and the conversational history thus far (see an example in Figure~\ref{fig:cqa}(a)). 
Formally, for the $i$-th turn, the human asks a question based on the previous conversation,

\begin{equation}
    \label{eq:human_q}
    \Q_i\sim \human(\Q_1, \A_1, ..., \Q_{i-1}, \A_{i-1}),
\end{equation}
and then the model answers it based on both the history and the passage,
\begin{equation}
    \label{eq:human_a}
    \A_i\sim \model(P, \Q_1, \A_1, ..., \Q_{i-1}, \A_{i-1}, \Q_i),
\end{equation}
where $\Q_i$ and $\A_i$ represent the question and the answer at the $i$-th turn.
If the question is unanswerable from $P$, we simply denote $\A_i$ as \unans{}.
The model $\model$ is then evaluated by the correctness of answers.

Evaluating {\convqa{}} systems requires human in the loop and is hence expensive.
Instead, current  benchmarks use {automatic evaluation} with {gold history} (\emph{\gold{}}) and collect a set of human-human conversations for automatic evaluation.
For each passage, 
one annotator asks questions without seeing the passage,
while the other annotator provides the answers. 
Denote the collected questions and answers as $\gcolor \Qg_i$ and $\gcolor \Ag_i$.
In gold-history evaluation,
the model is inquired with pre-collected questions $\gcolor \Qg_i$ and the gold answers as history:
\begin{equation}
    \label{eq:gold}
    \A_i\sim \model(P, {\gcolor \Qg_1}, {\gcolor \Ag_1}, ..., {\gcolor\Qg_{i-1}}, {\gcolor\Ag_{i-1}}, {\gcolor\Qg_i}),
\end{equation}
and we evaluate the model by comparing $A_i$ to $\gcolor \Ag_i$ (measured by word-level F1).
This process does not require human effort but cannot truly reflect the distribution of human-machine conversations,
because unlike human questioners who may ask different questions based on different model predictions, this static process ignores model predictions and always asks the pre-collected question. 

In this work, we choose the QuAC dataset~\cite{choi2018quac} as our primary evaluation because it is closer to real-world information-seeking conversations, where the questioner \emph{cannot} see the evidence passage during the dataset collection. It prevents the questioner asking questions that simply overlaps with the passage and encourages unanswerable questions.
%
QuAC also adopts \emph{extractive} question answering that restricts the answer as a span of text, which is generally considered easier to evaluate.


\subsection{Models}

For human evaluation and analysis, we choose the following four \convqa{} models with different model architectures and training strategies:

\paragraph{\bert{}.}
It is a simple BERT~\cite{devlin2019bert} baseline which concatenates the previous two turns of question-answer pairs, the question, and the passage as the input and predicts the answer span.\footnote{We use \texttt{bert-base-uncased} as the encoder.} 
This model is the same as the ``BERT + PHQA'' baseline in \citet{qu2019bert}.

\paragraph{\graphflow{}.}
\citet{chen2019graphflow} propose a recurrent graph neural network
on top of BERT embeddings to model the dependencies between the question, the history and the passage.

\paragraph{\ham{}.}
\citet{qu2019attentive} propose a history attention mechanism (HAM) to softly select the most relevant previous turns.

\paragraph{\excord{}.}
\citet{kim2021learn} train a question rewriting model on CANARD~\cite{elgohary2019unpack} to generate context-independent questions, and then use both the original and the generated questions to train the QA model. This model achieves the current state-of-the-art on QuAC (67.7\% F1).

For all the models except BERT, we use the original implementations for a direct comparison. 
We report their performance on both standard benchmark and our evaluation in Table~\ref{tab:performance-acc_f1}. 


%% file: sections/Human.tex

\section{Human Evaluation}
\label{sec:human}



\subsection{Conversation collection}
In this section, we carry out a large-scale human evaluation with the four models discussed above. We collect human-machine conversations using 100 passages from the QuAC development set on Amazon Mechanical Turk.\footnote{We restrict the annotators from English-speaking countries, and those who have finished at least 1,000 HITS with an acceptance rate of $>$95\%. The compensation rate for Amazon Mechanical Turk workers is calculated using \$15/h.}
We also design a set of qualification questions to make sure that the annotators fully understand our annotation guideline.
For each model and each passage, we collect three conversations from three different annotators.

We collect each conversation in two steps:

(1) The annotator has no access to the passage and asks questions. The model extracts the answer span from the passage or returns \unans{} in a human-machine conversation interface.\footnote{We used ParlAI~\cite{miller2017parlai} to build the interface.}
We provide the title, the section title, the background of the passage, and the first question from QuAC
as a prompt to annotators.
Annotators are required to ask at least 8 and at most 12 questions.
We encourage context-dependent questions, but also allow open questions like ``What else is interesting?'' if asking a follow-up question is difficult. (2) After the conversation ends, the annotator is shown the passage and asked to check whether the model predictions are correct or not.


We noticed that the annotators are biased when evaluating the correctness of answers. 
For questions to which the model answered \unans{}, annotators tend to mark the answer as incorrect without checking if the question is answerable.
Additionally, for answers with the correct types (e.g. a date as an answer to ``When was it?''), annotators tend to mark it as correct without verifying it from the passage. Therefore, we asked another group of annotators to verify question answerability and answer correctness.

\subsection{Answer validation}

For each collected conversation, we ask two additional annotators to validate the annotations.
First, each annotator reads the passage before seeing the conversation.
Then, the annotator sees the question (and question only) and selects whether the question is (a) ungrammatical, (b) unanswerable, or (c) answerable.
If the annotator chooses ``answerable'', the interface then reveals the answer and asks about its correctness.
If the answer is ``incorrect'', the annotator selects the correct answer span from the passage.
We discard all questions that both annotators find ``ungrammatical'' and  the correctness is taken as the majority of the 3 annotations.




\input{tables/human_stat}

\subsection{Annotation statistics}
\label{sec:anno_agreement}

In total, we collected 1,446 human-machine conversations and 15,059 question-answer pairs. We release this collection  as an important source that complements existing {\convqa{}} datasets. Numbers of conversations and question-answer pairs collected for each model are shown in Table~\ref{tab:human_collect}.
The data distribution of this collection is very different from the original QuAC dataset (human-human conversations): we see more open questions and unanswerable questions, due to less fluent conversation flow caused by model mistakes, and that models cannot provide feedback to questioner about whether an answer is worth following up like human answerers do
(more analysis in \S\ref{sec:obs1}).

Deciding the correctness of answers is challenging even for humans in some cases, especially when questions are short and ambiguous.
We measure annotators' agreement and calculate the
Fleiss' Kappa~\cite{fleiss1971measuring}
on the agreement between annotators in the validation phase.
We achieve $\kappa=0.598$ (moderate agreement) of overall annotation agreement. Focusing on answerability annotation, we have $\kappa=0.679$ (substantial agreement).






%% file: tables/human_stat.tex

\begin{table}[t]
    \begin{center}
    \resizebox{0.98\columnwidth}{!}{
    \begin{tabular}{l|c cc c c}
    \toprule

    & \multicolumn{4}{c}{\tf{Human Evaluation}} & \multirow{2}{*}{\tf{QuAC}}\\
     \cmidrule(lr){2-5}
      &  {\bert} &{GF} & {\ham} & {\excord} & \\
    \midrule
    \# C & 357& 359&373&357 & 1,000 \\
    \# Q  & 3,755 &  3,666 & 3,959 & 3,679  & 7,354\\
    \bottomrule
    \end{tabular}}
    \end{center}
    \caption{Number of conversations (\# C) and questions (\# Q) collected in human evaluation, using 100 passages from the QuAC development set. 
    We also add QuAC \ti{development} set for reference. GF: {\graphflow{}}.
    }
    \vspace{-5pt}
    \label{tab:human_collect}
    \end{table}

%% file: sections/Goldvshuman.tex

\section{Disagreements between Human and Gold-history Evaluation}
\label{sec:gold_vs_human}

We now compare the results from our human evaluation and gold-history (automatic) evaluation.
Note that the two sets of numbers are not directly comparable:
(1) the human evaluation reports accuracy, while the automatic evaluation reports F1 scores;
(2) the absolute numbers of human evaluation are much higher than those of automatic evaluations. 
For example, for the \bert{} model, the human evaluation accuracy is 82.6\% while the automatic evaluation F1  is only 63.2\%. 
The reason is that, in automatic evaluations, the gold answers cannot capture all possible correct answers to open-ended questions or questions with multiple answers; 
however, the human annotators can evaluate the correctness of answers easily.
Nevertheless, we can compare relative rankings between different models.

Figure~\ref{fig:gold_vs_human} shows different trends between human evaluation and gold-history evaluation (\gold{}). Current standard evaluation cannot reflect model performance in human-machine conversations:
(1) Human evaluation and \gold{} rank \bert{} and \graphflow{} differently; 
especially, \graphflow{} performs much better in automatic evaluation, but worse in human evaluation. 
(2) The gap between \ham{} and \excord{} is significant (F1 of 65.4\% vs 67.7\%)
in the automatic evaluation but the two models perform similarly in human evaluation (accuracy of 87.8\% vs 87.9\%).


\input{tables/gold_vs_human}

%% file: tables/gold_vs_human.tex

\begin{figure}
    \centering
    \includegraphics[width=0.9\columnwidth]{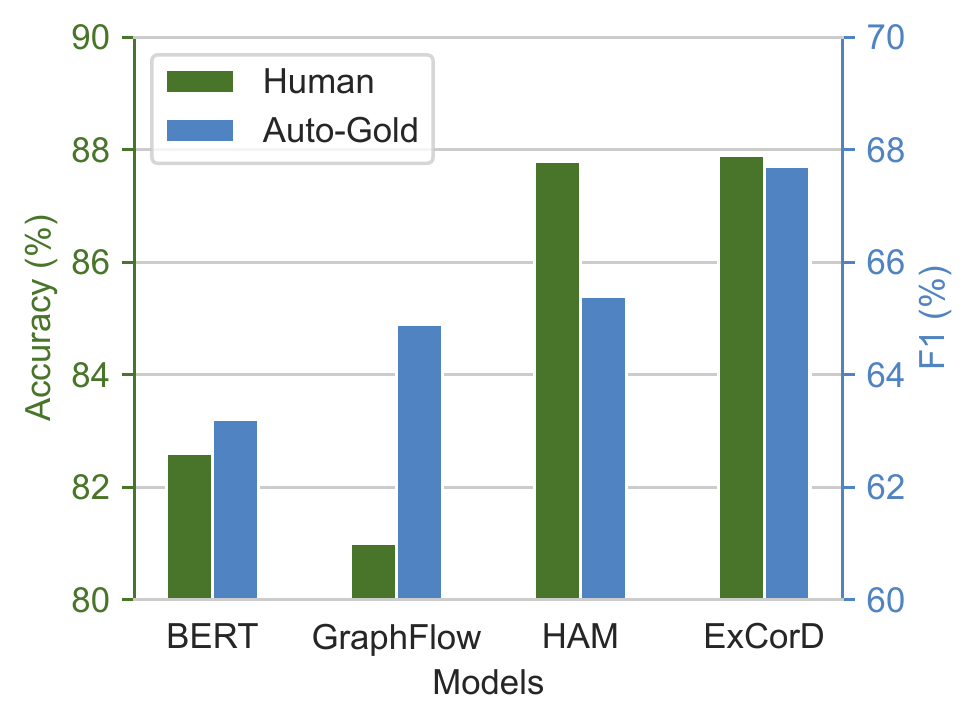}
    \caption{Model performance of human evaluation (accuracy, left) and Auto-Gold (F1, right). 
    Scales for accuracy and F1 are different. 
    Human evaluation and \gold{} rank \bert{} and \graphflow{} differently.
    }
    \vspace{-5pt}
    \label{fig:gold_vs_human}
\end{figure}


%% file: sections/Pred.tex

\section{Strategies for Automatic Evaluation}
\label{sec:pred}

The inconsistency between human evaluation and gold-history evaluation suggests that we need better ways to evaluate and develop our \convqa{} models.
When being deployed in realistic scenarios, the models would never have access to the ground truth (gold answers) in previous turns and are only exposed to the conversational history and the passage.
Intuitively, we can simply replace gold answers by the predicted answers of models and we name this as \tf{predicted-history} evaluation  (\emph{\pred{}}).
Formally, the model makes predictions based on the questions and its own answers:
\vspace{-0.5em}
\begin{equation}
    \label{eq:pred}
    \A_i\sim \model(P, {\gcolor\Qg_1}, \A_1, ..., {\gcolor\Qg_{i-1}}, \A_{i-1}, {\gcolor\Qg_i}).
\end{equation}
This evaluation has been suggested by several recent works~\cite{mandya-etal-2020-history,siblini-etal-2021-towards}, which reported a significant performance drop using predicted history.
We observe the same performance degradation, shown in Table~\ref{tab:performance-acc_f1}.

However, another issue naturally arises with predicted history:  $\gcolor \Qg_{i}$s were written by the dataset annotators based on (${\gcolor\Qg_1}, {\gcolor\Ag_1}, ..., {\gcolor\Qg_{i-1}}, {\gcolor\Ag_{i-1}}$), which may become unnatural or invalid when the history is changed to (${\gcolor\Qg_1}, \A_1, ..., {\gcolor\Qg_{i-1}}, \A_{i-1}$).



\input{tables/exp_of_invalid}

\subsection{Predicted history invalidates questions}
\label{sec:invalid_q}

We examined 100 QuAC conversations with the best-performing model (\excord{}) and identified three categories of invalid questions caused by predicted history.
We find that 23\% of the questions become invalid after using the predicted history.
We summarize the types of invalid questions as follows (see detailed examples in Figure~\ref{tab:exp_of_invalid}):


\begin{itemize}[leftmargin=*]
   \setlength\itemsep{-0.2em}
    \item \tf{Unresolved coreference}  (44.0\%). The question becomes invalid for containing  either a pronoun or a definite noun phrase 
    that refers to an entity unresolvable without the gold history.

    \item \tf{Incoherence} (39.1\%). The question is incoherent with the conversation flow (e.g., mentioning an entity non-existent in predicted history). While humans may still answer the question using the passage, this  leads to an unnatural conversation and a train-test discrepancy for models.


    \item \tf{Correct answer changed} (16.9\%). 
    The answer to this question with the predicted history changes from when it is based on the gold history.


\end{itemize}


We further analyze the reasons for the biggest ``unresolved coreference'' category and find that the model either gives an incorrect answer to the previous question (``incorrect prediction'', 39.8\%), or the model predicts a different (yet correct) answer to an open question (``open question'', 37.0\%), or the model returns \unans{} incorrectly (``no prediction'', 9.5\%), or the gold answer is longer than  prediction and the next question depends on the extra part (``extra gold information'', 13.6\%).  Invalid questions result in compounding errors, which may further affect how the model interprets the following questions. 

%% file: tables/exp_of_invalid.tex

\begin{figure}[t]
    \centering
    \resizebox{0.97\columnwidth}{!}
    {\begin{tabular}{cp{\columnwidth}}
    \toprule
        \multicolumn{2}{l}{\tf{Unresolved coreference} (44.0\%)}\\
    \midrule
        $\Qg_1$: & What was Frenzal Rhomb's first song?\\
        $\Ag_1$: & {Punch in the Face}.\\
        $\A_1$: & \unans{}. \\
        ~\vspace{-8pt}\\
        $\Qg_2$: & How did \ti{\color{purple}{it}} fare?\\

    \midrule
        \multicolumn{2}{l}{\tf{Incoherence} (39.1\%)}\\
    \midrule

        $\Qg_1$: & Did Billy Graham succeed in becoming a chaplain? \\
        $\Ag_1$: & {He \ti{\color{purple}contracted mumps} shortly after...}\\
        $\A_1$: & After a period of recuperation in Florida, he ...\\
        ~\vspace{-8pt}\\
        $\Qg_2$: & Did he retire after his \ti{\color{purple}mumps diagnosis}? \\

    \midrule
    \multicolumn{2}{l}{\tf{Correct answer changed} (16.9\%)}\\
    \midrule

        $\Qg_1$: & Are there any other interesting aspects? \\
        $\Ag_1$: & ... \ti{\color{purple}Steve Di Giorgio returned to the band}... \\
        $\A_1$: & ... bassist Greg Christian had left Testament again...\\
        ~\vspace{-8pt}\\
        $\Qg_2$: & What happened following \ti{\color{purple}this change in crew}?\\



    \bottomrule
    \end{tabular}
    }
    \caption{
    Examples of invalid questions with predicted history. Some are shortened for better demonstration. $\Qg_i, \Ag_i$: questions and gold answers from the collected dataset, $\A_i$: model predictions.
    }
    \label{tab:exp_of_invalid}
\end{figure}

%% file: sections/Automatic.tex
\subsection{Evaluation with question substitution}

\begin{figure}[t]
  \centering
  \includegraphics[width=0.98\columnwidth]{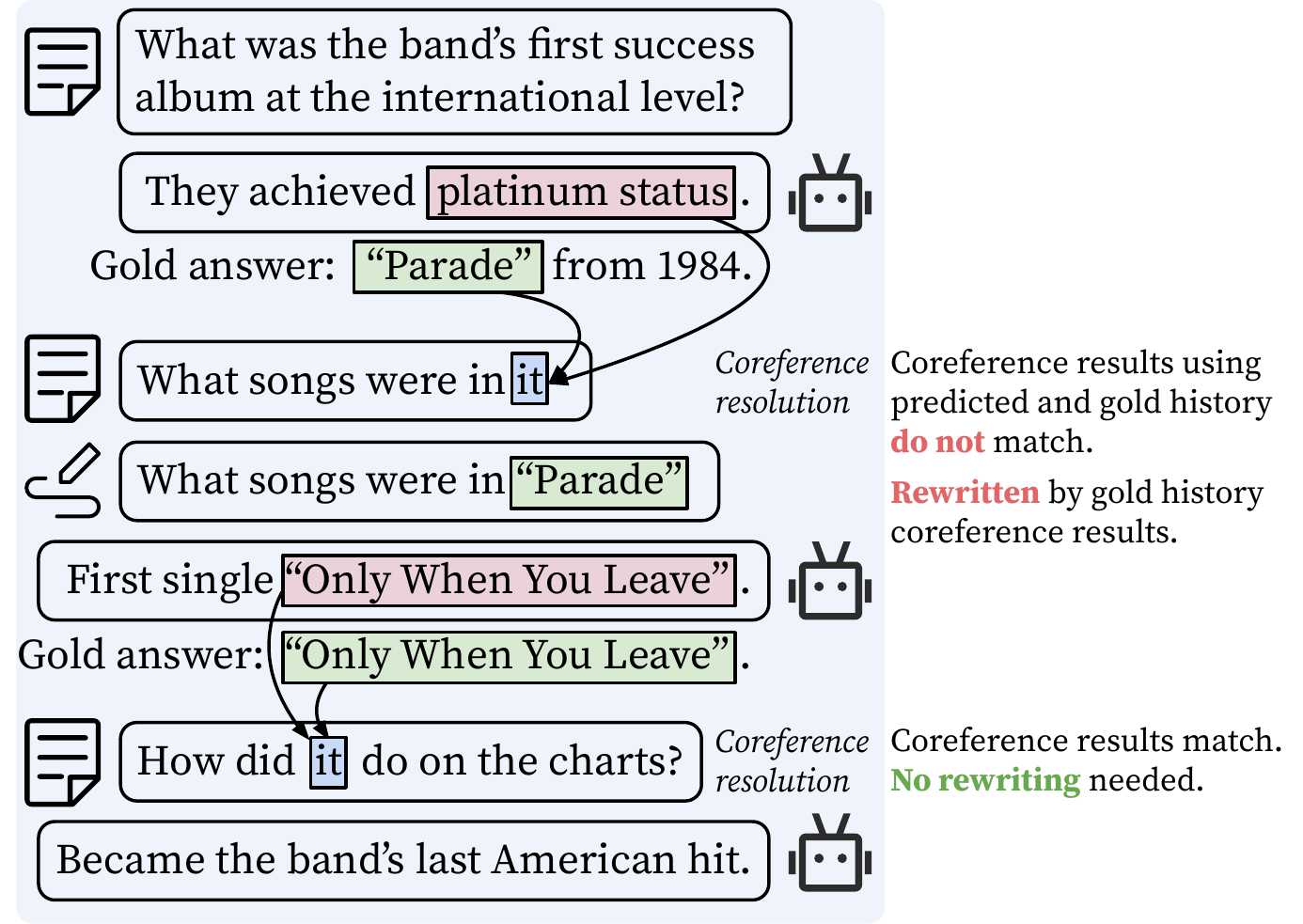}
  \caption{
  An example of question rewriting. 
  We rewrite the second question with referent in the gold history, because predicted and gold history have  different coreference results.
  We do not rewrite the third question as coreference results are the same.
  }
  \label{fig:rewrite}
\end{figure}

\input{tables/overall_f1_acc}

\label{sec:method_detect}
Among all the invalid question categories, ``unresolved coreference'' questions are the most critical ones. They lead to incorrect interpretations of questions and hence wrong answers. We propose to improve our evaluation by first detecting these questions using a state-of-the-art coreference resolution system~\cite{lee2018higher}\footnote{We use the coreference model from {AllenNLP}~\cite{gardner2018allennlp}.}, and then substituting them with either 
rewriting the questions in-place and replacing the questions with their context-independent counterparts.




\paragraph{Detecting invalid questions.} 
We make the assumption that if the coreference model resolves mentions in $\Qg_i$ differently between using gold history $({\gcolor\Qg_1}, {\gcolor\Ag_1}, ..., {\gcolor\Ag_{i-1}}, {\gcolor\Qg_i})$ and predicted history $({\gcolor\Qg_1}, \A_1, ..., \A_{i-1}, {\gcolor\Qg_i})$, then ${\gcolor \Qg_i}$ is identified as having an unresolved coreference issue.

The inputs to the coreference model for ${\Qg_i}$ are the following:
\begin{equation*}
  \resizebox{.98\hsize}{!}{%
  $
  \begin{aligned}
    S^*_i &= [\bg{}; {\gcolor\Qg_{i-k}}; {\gcolor A^*_{i-k}}; {\gcolor\Qg_{i-k+1}}; {\gcolor A^*_{i-k+1}}; ...; {\gcolor\Qg_i}]\\
    S_i &= [\bg{}; {\gcolor\Qg_{i-k}}; A_{i-k}; {\gcolor\Qg_{i-k+1}}; A_{i-k+1}; ...; {\gcolor\Qg_i}],
  \end{aligned}
  $
  }
\end{equation*}
where $\bg{}$ is the background,
 $S^*_i$ and $S_i$ denote the inputs for gold and predicted history.
After the coreference model returns entity cluster information given $S^*_i$ and $S_i$, we extract a list of entities $E^*=\{e^*_1,...,e^*_{|E^*|}\}$ and $E=\{e_1,...,e_{|E|}\}$.\footnote{We are only interested in the entities mentioned in the current question $ \Qg_i$ and we
filter out named entities (e.g., the \emph{National Football League}) because they can be understood without coreference resolution.
}
We say $ \Qg_i$ is \emph{valid} only if $E^*=E$, that is,
\begin{equation*}
|E^*|=|E| \text{~and~} e^*_j=e_j, \forall e_j\in E,
\end{equation*}
assuming $e^*_j$ and $e_j$ have a shared mention in $\gcolor \Qg_i$.
We determine whether $e^*_j=e_j$ by checking if $\text{F1}(s^*_{j}, s_{j}) > 0$,
where $s^*_{j}$ and $s_{j}$ are the \ti{first} mention of $e^*_j$ and $e_j$ respectively, and $\text{F1}$ is the word-level F1 score, i.e., $e^*_j=e_j$ as long as their first mentions have word overlap.
The reason we take the F1 instead of exact match to check whether the entities are the same is stated in Appendix \ref{app:detector}. 

%


\paragraph{Question rewriting through entity substitution.}
Our first strategy is to substitute the entity names  in $Q_i^*$ with entities in $E^*$, if $Q_i^*$ is invalid.
The rewritten question, instead of the original one, will be used in the conversation history and fed into the model. 
We denote this evaluation method as \tf{rewritten-question} evaluation (\emph{\rewrite{}}), and Figure~\ref{fig:rewrite} illustrates a concrete example.

To analyze how well \rewrite{} does in detecting and rewriting questions, we manually check 100 conversations of \excord{} from the QuAC development set.
We find that \rewrite{} can detect invalid questions with a precision of 72\% and a recall of 72\% (more detailed analysis in Appendix~\ref{app:rewrite_quality}).
An example of correctly detected and rewritten question is presented in Figure~\ref{fig:rewrite}. 



\paragraph{Question replacement using CANARD.}
Another strategy is to replace the invalid questions with  context-independent questions. 
The CANARD dataset~\cite{elgohary2019unpack} provides such a resource, which contains human-rewritten context-independent version of QuAC's questions. 
Recent works~\cite{anantha-etal-2021-open, elgohary2019unpack} have proposed training sequence-to-sequence models on such dataset
to rewrite questions; 
however, since the performance of the question-rewriting models is upper bounded by the human-rewritten version, 
we simply use CANARD for question replacement. 
We denote this strategy as \tf{replaced-question} evaluation (\emph{\replace{}}).
Because collecting context-independent questions is expensive, \replace{} is limited to evaluating models on QuAC; 
it is also possible to be extended to other datasets by training a question rewriting model, as demonstrated in existing work. 




%% file: tables/overall_f1_acc.tex
\begin{table*}[t]
\begin{center}
    \resizebox{0.94\linewidth}{!}{
\begin{tabular}{lcccccccc}
\toprule
 & \multicolumn{4}{c}{\tf{All}} &  \multicolumn{4}{c}{\tf{Answerable questions}} \\
 \cmidrule(lr){2-5}
 \cmidrule(lr){6-9}
 & {\bert} &{GraphFlow} & {\ham} & {\excord} &  {\bert}& {GraphFlow}  & {\ham} & {\excord} \\
\midrule
Auto-Gold (F1)    & 63.2  & 64.9  & 65.4  & 67.7  & 61.8 & 66.6  & 64.5 & 66.4 \\
Auto-Pred (F1)    & 54.6  & 49.6  & 57.2  & 61.2  & 52.7 & 54.5  & 54.6 & 59.2 \\
Auto-Rewrite (F1) & 54.5  & 48.2  & 57.3  & 61.9  & 51.2 & 51.9  & 55.1 & 59.7  \\
Auto-Replace (F1) & 54.2  & 47.8  & 57.1  & 61.7  & 50.7 & 51.7  & 54.8 & 59.7  \\
\midrule
\cellcolor{cid}Human (Accuracy)  & \cellcolor{cid}82.6  & \cellcolor{cid}81.0 &  \cellcolor{cid}87.8 & \cellcolor{cid}87.9 & \cellcolor{cid}75.9 & \cellcolor{cid}83.2  & \cellcolor{cid}84.8 & \cellcolor{cid}85.3  \\
\bottomrule
\end{tabular}}
\end{center}
\caption{
Model performance in automatic and human evaluations.
We report \emph{overall performance} on all questions and also performance on \emph{answerable questions} only.
}
\label{tab:performance-acc_f1}
\vspace{-0.8em}
\end{table*}

%% file: sections/Experiment.tex

\section{Automatic vs Human Evaluation}
\label{sec:exp}

%
%
%
%

In this section, we compare human evaluation  with all the automatic evaluations we have introduced: gold-history (\gold{}), predicted-history  (\pred{}), and our proposed \rewrite{} and \replace{} evaluations.
We first explain the metrics we use in the comparison (\S\ref{sec:metric}) and then discuss the findings (\S\ref{sec:obs1} and \S\ref{sec:obs2}).



\subsection{Agreement metrics}
\label{sec:metric}

\paragraph{Model performance and rankings.}
We first consider using model performance reported by different evaluation methods.
Considering numbers of automatic and human evaluations are not directly comparable,
we also calculate models' rankings
and compare whether the rankings are consistent between automatic  and human evaluations.
Model performance is reported in
Table~\ref{tab:performance-acc_f1}.
In human evaluation,
\graphflow{} $<$ \bert{} $<$ \ham{} $\approx$ \excord{};
in \gold{},
\bert{} $<$ \graphflow{} $<$ \ham{} $<$ \excord{};
in other automatic evaluations,
\graphflow{} $<$ \bert{} $<$ \ham{} $<$ \excord{}.

\paragraph{Statistics of unanswerable questions.}
Percentage of unanswerable questions is an important aspect in conversations.
Automatic evaluations using static datasets have a fixed number of unanswerable questions, while in human evaluation, the percentage of unanswerable questions asked by human annotators varies with different models.
The statistics of unanswerable questions is shown in
Table~\ref{tab:no_ans}.

\input{tables/num_unans}

\input{tables/rank_align.tex}

\input{tables/unanswerable}

\paragraph{Pairwise agreement.}
For a more fine-grained evaluation,
we perform a passage-level comparison for every pair of models.
More specifically, for every single passage we use one automatic metric to decide whether model $A$ outperforms model $B$ (or vice versa) and examine the percentage of passages that the automatic metric agrees with human evaluation.
For example, if the pairwise agreement of \bert{}/\excord{} between human evaluation and \gold{} is $52\%$, it means that \gold{} and human evaluation agree on $52\%$ passages in terms of which model is better.
Higher agreement means the automatic evaluation is closer to human evaluation. 
Figure~\ref{fig:pair_align} shows the results of pairwise agreement.

\subsection{Automatic evaluations have a significant distribution shift from human evaluation}
\label{sec:obs1}


We found that automatic evaluations have a significant distribution shift from human evaluation. 
We draw this conclusion from the  following points.

\begin{itemize}[leftmargin=*]
\item
Human evaluation shows a much higher model performance than all automatic evaluations, as shown in Table~\ref{tab:performance-acc_f1}.
Two reasons may cause this large discrepancy:
(a) Many \convqa{} questions have multiple possible answers, and it is hard for the static dataset in automatic evaluations to capture all the answers. 
It is not an issue in human evaluation because all answers are judged by human evaluators.
(b) There are more unanswerable questions and open questions in human evaluation (reason discussed in the next paragraph), which are easier---for example, models are almost always correct when answering questions like ``What else is interesting?''. 





\item Human evaluation has a much higher unanswerable question rate, as shown in Table~\ref{tab:no_ans}.
The reason is that in human-human data collection, the answers are usually correct and the questioners can ask followup questions upon the high-quality conversation;
in human-machine interactions, since the models can make mistakes, the conversation flow is less fluent
and it is harder to have followup questions. Thus,
questioners chatting with models tend to ask more open or unanswerable questions.

\item All automatic evaluation methods have a pairwise agreement lower than 70\% with human evaluation, as shown in Figure~\ref{fig:gold_vs_human}. 
This suggests that all automatic evaluations cannot faithfully reflect the model performance of human evaluation.

\end{itemize}


\subsection{\rewrite{} is closer to human evaluation}
\label{sec:obs2}
First, we can clearly see that among all automatic evaluations, \gold{}
deviates the most from the human evaluation.
From Table~\ref{tab:performance-acc_f1}, only \gold{} shows different rankings from human evaluation, while \pred{}, \rewrite{}, and \replace{} show consistent rankings to human judgments.

In Figure~\ref{fig:gold_vs_human},
we see that \gold{} has the lowest agreement with human evaluation;
among others, \rewrite{} better agrees with human evaluation for most model pairs.
Surprisingly, \rewrite{} is even better than \replace{}---which uses human-written context independent questions---in most cases.
After checking the Auto-Replace conversations, we found that human-written context independent questions are usually much longer than  QuAC questions and introduce extra information into the context, which leads to out-of-domain challenges for \convqa{} models (example in Appendix~\ref{app:replace_error}).
It shows that our rewriting strategy can better reflect real-world performance of conversational QA systems.
However, \rewrite{} is not perfect---we see that when comparing G/E or G/H, \pred{} is better than \rewrite{}; in all model pairs, the agreement between human evaluation and \rewrite{} is still lower than 70\%. 
This calls for further effort in designing better automatic evaluation in the future.

%% file: tables/num_unans.tex

\begin{table}[t]
    \begin{center}
        \resizebox{0.9\columnwidth}{!}{
    \begin{tabular}{ccccc}
    \toprule
    \multicolumn{4}{c}{\tf{Human Evaluation}} & \multirow{2}{*}{\tf{QuAC}}\\
    \cmidrule(lr){1-4}
      {\bert} &{GF} & {\ham} & {\excord} & \\
    \midrule
    34.6 & 20.6 & 34.1 & 33.2 & 20.2\\
    \bottomrule
    \end{tabular}}
    \end{center}

    \caption{
        Percentage of unanswerable questions in human evaluation (it varies with different models) and the original QuAC dataset (used in all automatic evaluations). GF: \graphflow{}.
    }
    \vspace{-10pt}
    \label{tab:no_ans}
    \end{table}


%% file: tables/rank_align.tex

\begin{figure}[t]
    \centering
    \includegraphics[width=\linewidth]{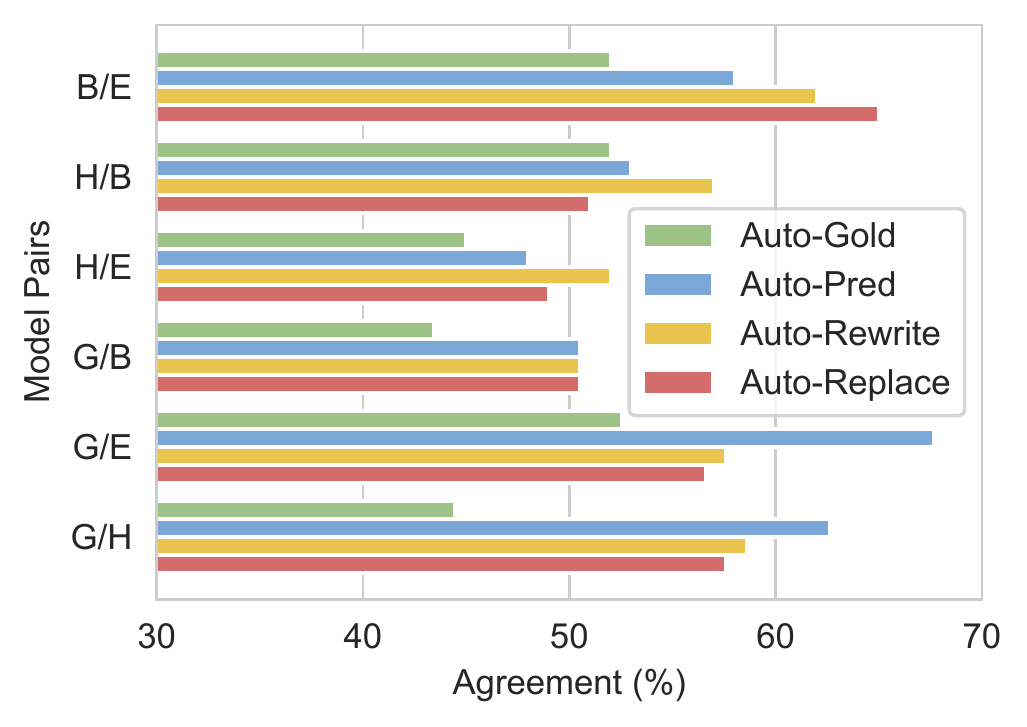}
    \caption{Pairwise agreement of different model pairs comparing automatic evaluations to human evaluation. B: \bert{}; G: \graphflow{}; H: \ham{}; E: \excord{}. 
    }
    \vspace{-10pt}
    \label{fig:pair_align}
\end{figure}

%% file: tables/unanswerable.tex

\begin{table*}[t]
\begin{center}
    \resizebox{0.9\textwidth}{!}{
\begin{tabular}{lcccccccccccc}
\toprule
\cmidrule(lr){2-13}
& \multicolumn{4}{c}{\tf{Predicted unanswerable Q.}} & \multicolumn{4}{c}{\tf{Precision}} & \multicolumn{4}{c}{\tf{Recall}} \\
\cmidrule(lr){2-5}
\cmidrule(lr){6-9}
\cmidrule(lr){10-13}
& B & G & H & E &  B& G  & H & E &  B& G  & H & E \\
\midrule
\gold{} & 27.1 & 21.5 & 27.1 & 28.3  & 56.8 & 62.3 & 57.1 & 57.9 & 68.1 & 59.3 & 68.4 & 72.5  \\
\pred{} & 27.8 & 13.8 & 28.6 & 28.9 &  50.0 & 53.9 & 52.3 & 53.3 & 61.4 & 33.0 & 66.1 & 68.2 \\
\rewrite{} & 27.3 & 13.1 & 25.1 & 26.0  & 48.6 & 55.0 & 52.4 & 53.9 &  65.7 & 35.7 & 65.1 & 69.4\\
\replace{} & 27.5 & 12.9 & 25.2 & 25.7  & 48.6 & 54.2 & 52.1 & 53.8 & 66.1 & 34.7 & 64.9 & 68.4 \\
\midrule
\cellcolor{cid}Human  & \cellcolor{cid}42.3 & \cellcolor{cid}14.7 & \cellcolor{cid}37.2 & \cellcolor{cid}36.0 & \cellcolor{cid}75.0 & \cellcolor{cid}93.0 & \cellcolor{cid}86.8 & \cellcolor{cid}87.4  & \cellcolor{cid}95.2 & \cellcolor{cid}72.5 & \cellcolor{cid}93.7 & \cellcolor{cid}93.3 \\

\bottomrule
\end{tabular}}
\end{center}
\caption{
    The percentage of models' predicted unanswerable questions, and the precision and recall for detecting unanswerable questions in different evaluations. 
    B: \bert{}; G: \graphflow{}; H: \ham{}; E: \excord{}.
}
 \vspace{-3pt}
\label{tab:performance-unans}
\end{table*}

%% file: sections/ModelAnalysis.tex
\input{tables/model_structure}

\input{tables/exp_of_context_modeling}

\section{Towards Better Conversational QA}

\label{sec:analysis}





With insights drawn from human evaluation and comparison with automatic evaluations,
we discuss the impact of different modeling strategies, as well as future directions towards building better conversational question answering systems.



%

\paragraph{Modeling question dependencies on conversational context.}
When we focus on \ti{answerable questions} (Table~\ref{tab:performance-acc_f1}),
we notice that \graphflow{}, \ham{} and \excord{} 
perform much better than \bert{}.
We compare the modeling differences of the four systems in Figure~\ref{fig:model_structures},
and identify that  all the three better systems explicitly model
the question dependencies on the conversation history and the passage:
both \graphflow{} and \ham{} highlight repeated mentions in questions and conversation history by special embeddings (turn marker and PosHAE) and use attention mechanism to select the most relevant part from the context;
\excord{} adopts a question rewriting module that generates context-independent questions given the history and passage.
All those designs help models better understand the question in a conversational context.
Figure~\ref{tab:exp_of_context_modeling} gives an example where \graphflow, \ham{} and \excord{}  resolved the question from long conversation history while \bert{} failed.

\paragraph{Unanswerable question detection.}
Table~\ref{tab:performance-unans} demonstrates models' performance in detecting \emph{unanswerable questions}.
We notice that \graphflow{} predicts much fewer unanswerable questions than the other three models, and has a high precision and a low recall in unanswerable detection. This is because \graphflow{} uses a separate network for predicting unanswerable questions, which is harder to calibrate, while the other models jointly predict unanswerable questions and answer spans.


This behavior has two effects:
(a) \graphflow{}'s overall performance is dragged down by its poor unanswerable detection result (Table~\ref{tab:performance-acc_f1}).
(b) In human evaluation, annotators ask fewer unanswerable questions with \graphflow{} (Table~\ref{tab:no_ans})---when the model outputs more, regardless of correctness, the human questioner has a higher chance to ask passage-related followup questions.
Both suggest that how well the model detects unanswerable questions significantly affects its performance and the flow in human-machine conversations.

\paragraph{Optimizing towards the new testing protocols.}
Most existing works on \convqa{} modeling focus on optimizing towards \gold{} evaluation.
Since \gold{} has a large gap from the real-world evaluation,
more efforts are needed in optimizing towards the human evaluation, or \rewrite{}, which better reflects human evaluation.
One potential direction is to improve models' robustness given noisy conversation history, which simulates the inaccurate history in real world that consists of models' own predictions.
In fact, prior works \cite{mandya-etal-2020-history, siblini-etal-2021-towards} that used predicted history in training showed that it benefits the models in predicted-history evaluation.



%% file: tables/model_structure.tex
\begin{figure*}[t]
    \centering
    \includegraphics[width=\textwidth]{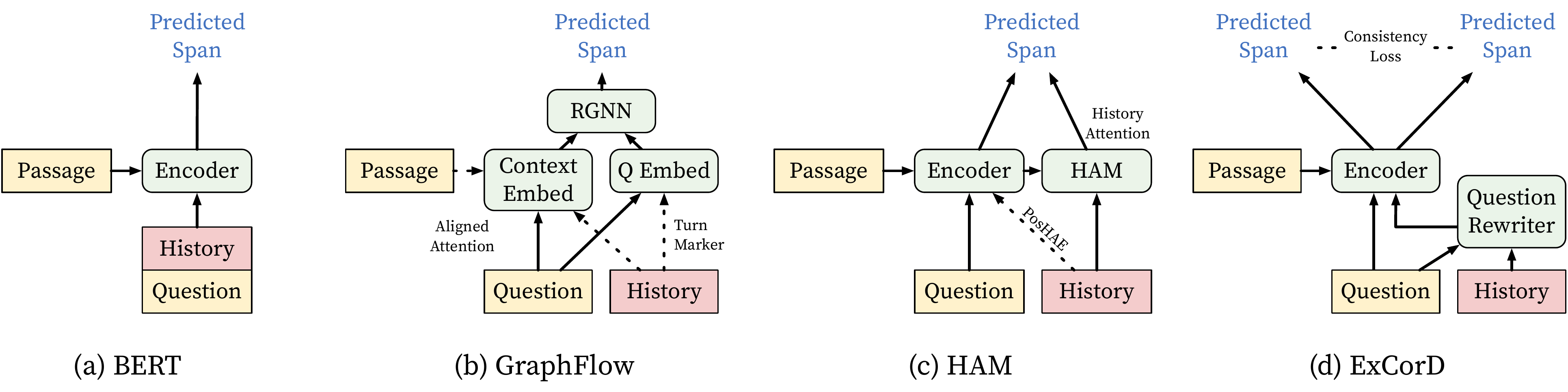}
    \caption{Modeling structures of \bert{}, \graphflow{}, \ham{}, and \excord{}.
    }
    \vspace{-5pt}
    \label{fig:model_structures}
\end{figure*}

%% file: tables/exp_of_context_modeling.tex

\begin{figure}[t]
    \centering
    \resizebox{0.98\columnwidth}{!}
    {\begin{tabular}{llc}
    \toprule

       \multicolumn{3}{l}{\tf{Tom McCall -- Vortex I}}\\
       \multicolumn{3}{l}{... McCall decided to hold a rock festival at Milo McIver }\\
       \multicolumn{3}{l}{State Park, Oregon called ``Vortex I: A Biodegradable }\\
       \multicolumn{3}{l}{Festival of Life''...}\\

    \midrule
        $Q^*_1$: & Was Vortex I popular? & \\
        B: & \emph{\color{purple}The festival}, ``The Governor's Pot Party'' ... & \checkmark\\
        G/H/E: & \emph{\color{purple}The festival}, ``The Governor's Pot Party'' ...  & \checkmark\\
        ...\\
        $Q^*_4$: & Who played at \emph{\color{purple}the festival}? & \\
        B: & \unans{} & \xmark\\
        G/H/E: & Gold, The Portland Zoo, Osceola, Fox... & \checkmark\\
    \bottomrule
    \end{tabular}
    }
    \caption{
    An example of \bert{} failing to resolve  \emph{the festival} in $\Qg_4$, while all other models with explicit dependency modelings succeeded.
    }
    \label{tab:exp_of_context_modeling}
    \vspace{-0.8em}
\end{figure}

%% file: sections/Related.tex

\section{Related Work}
\label{sec:related}

%

\paragraph{Conversational question answering.}
In recent years, several conversational question answering datasets have emerged, such as QuAC~\cite{choi2018quac}, CoQA~\cite{reddy2019coqa}, and DoQA~\cite{campos-etal-2020-doqa}, as well as a few recent works focusing on conversational open-domain question answering~\cite{adlakha2021topiocqa,anantha-etal-2021-open,qu2020openretrieval}
Different from single-turn QA datasets~\cite{rajpurkar2016squad}, \convqa{} requires the model to understand the question in the context of conversational history.
There have been many methods proposed to improve \convqa{} performance~\cite{ ohsugi2019simple,chen2019graphflow,  qu2019attentive, kim2021learn} and significant improvements have been made on \convqa{} benchmarks.
Besides text-based \convqa{} tasks, there also exist \convqa{} benchmarks that require external knowledge or other modalities~\cite{saeidi2018interpretation,saha2018complex, guo2018dialog,visdial}.

Only recently has it been noticed that the current method of evaluating \convqa{} models is flawed. \citet{mandya-etal-2020-history, siblini-etal-2021-towards} point out that using gold answers in history
is not consistent with real-world scenarios
and propose to use predicted history for evaluation.
Different from prior works, in this paper, we conduct a large scale human evaluation to provide evidence for why gold-history evaluation is sub-optimal. In addition, we point out that even predicted-history evaluation has issues with invalid questions, for which we propose rewriting questions to further mitigate the gap.

\paragraph{Automatic evaluation of dialogue systems.}
Automatically evaluating dialogue systems is difficult due to the nature of conversations. 
In recent years, the NLP community has cautiously re-evaluated and identified flaws in many popular automated evaluation strategies of dialogue systems~\cite{liu2016not,sai2019re}, 
and have proposed new evaluation protocols to align more with human evaluation in a real-world setting: 
\citet{huang-etal-2020-grade,ye-etal-2021-towards-quantifiable} evaluate the coherence of the dialogue systems; 
\citet{gupta-etal-2019-investigating} explore to use multiple references for evaluation; 
\citet{mehri-eskenazi-2020-usr} propose an unsupervised and reference-free evaluation; 
\citet{lowe-etal-2017-towards,tao2018ruber,ghazarian2019better,shimanaka2019machine,sai-etal-2020-improving} 
train models to predict the relatedness score between references and model outputs, which are shown to be better than BLEU~\cite{papineni2002bleu} or ROGUE~\cite{lin-2004-rouge}.






%% file: sections/Conclusion.tex

\section{Conclusion}\label{sec:conclusion}

In this work, we carry out the first large-scale human evaluation on \convqa{} systems. 
We show that current standard automatic evaluation with gold history cannot reflect models' performance in human evaluation, and that human-machine conversations have a large distribution shift from static \convqa{} datasets of human-human conversations. 
To tackle these problems, we propose to use predicted history with rewriting invalid questions for evaluation, 
which reduces the gap between automatic evaluations and real-world human evaluation.
Based on the insights from the human evaluation results, we also nalyze current \convqa{} systems and identify promising directions for future development. 

%% file: sections/Appendix.tex

%

%
%
%
%



 

\section{Invalid Question Detection}
\label{app:detector}

In question rewriting, we use F1 instead of exact match to check whether two entites are the same. The reason is that sometimes the prediction may mention the same entity as the gold answer does, but with different names. Figure~\ref{tab:coref_ex} gives an example. 
Thus to avoid the false positive of detecting invalid questions, we take the F1 metric.

\input{tables/coref_ex}

\section{Quality of Rewriting Questions}
\label{app:rewrite_quality}


\paragraph{Detection.} After manually checking 100 conversations of \excord{} from the QuAC development set, we find that \rewrite{} can detect invalid questions with a precision of 72\% and a recall of 72\%. 
We notice that the coreference model sometimes detects the pronoun of the main character in the passage as insolvable, although it almost shows up in every question. 
This issue causes the low precision but is not a serious problem in our case -- 
whether rewriting the pronoun of the main character does not affect models' prediction much, 
because the model always sees the passage and knows who the main character is. 


\paragraph{Rewriting.} Among all correctly detected invalid questions, we further check the quality of rewriting, and in 68\% of the times \rewrite{} gives a correct context-independent questions. 
The most common error is being ungrammatical. 
For example, using the gold history of "... Dee Dee claimed that Spector once \tf{\ti{pulled}} a gun on him", the original question "Did they arrest him for doing \ti{this}?" was rewritten to "Did they arrest Phillip Harvey Spector for doing \tf{\ti{pulled}}?"
While this creates a distribution shift on question formats, 
it is still better than putting an invalid question in the flow. 

\section{Issue with Context Independent Questions}
\label{app:replace_error}

Figure~\ref{tab:replace_error} shows an example where extra information in context-independent questions confuses the model and leads to incorrect prediction.

\input{tables/replace_error}

%% file: tables/coref_ex.tex

\begin{figure}[ht]
\resizebox{0.98\columnwidth}{!}{
\begin{tabular}{p{\columnwidth}}
\toprule
    $Q^*_1$: Who is at the door?\\
    $A_1^*$: \ti{\color{purple} An elderly Chinese lady} and a little boy\\
    $A_1$: \ti{\color{purple} elderly Chinese lady}\\
    \vspace{1pt}
    $Q^*_2$: Is \ti{\color{purple} she} carrying something?\\
\bottomrule
\end{tabular}}
\caption{An example that the prediction may mention the same entity as the gold answer does with slightly different names.}
\label{tab:coref_ex}
\vspace{-1em}
\end{figure}

%% file: tables/replace_error.tex

\begin{figure}[t]
    \centering
    \resizebox{0.98\columnwidth}{!}
    {\begin{tabular}{ll}
    \toprule
        $Q^*$: & Did he go on to any other notable matches? \\
        ~\vspace{-8pt}\\
        $Q^W$: & Did \ti{\color{purple} he} go on to any other notable matches? \\
        $A^W$: & During the Test match series against Australia \\
        & in 2010, at the Melbourne Cricket Ground...\\
        ~\vspace{-8pt}\\
        $Q^P$: & Did \ti{\color{purple}Mohammad Amir} go on to any other \\ & notable matches, \ti{\color{orange} besides on 9 November 2009}? \\
        $A^P$: & Later in 2009, Pakistan toured Sri Lanka \\
    \bottomrule
    \end{tabular}
    }
    \caption{
    The context-independent question $Q^P$ by \Replace{} contains {\color{orange} extra information} that confuses the model. The rewritten question $Q^W$ did not change the original question and led to a correct answer.
    }
    \label{tab:replace_error}
\end{figure}